\documentclass[runningheads]{llncs}

\usepackage{eccv}

\usepackage{amsfonts}       
\usepackage{booktabs}       
\usepackage{graphicx}
\usepackage[misc]{ifsym}    
\usepackage{multirow}
\usepackage{tabularx}
\usepackage{tipa}           

\usepackage[pagebackref,breaklinks,colorlinks,citecolor=eccvblue]{hyperref}

\usepackage{orcidlink}

\newcommand{\OM}{HSImul3R\xspace}
\newcommand{\eg}{\emph{e.g}}

\newcolumntype{Y}{>{\centering\arraybackslash}X}
\renewcommand{\paragraph}[1]{%
  \par\medskip
  \noindent\textbf{#1.\ }%
}

\begin{document}

\title{\OM: Physics-in-the-Loop Reconstruction of Simulation-Ready Human–Scene Interactions}

\author{%
Yukang Cao\textsuperscript{1} \and
Haozhe Xie\textsuperscript{1} \and
Fangzhou Hong\textsuperscript{1} \and
Long Zhuo\textsuperscript{3} \and\\
Zhaoxi Chen\textsuperscript{1} \and
Liang Pan\textsuperscript{2}\and
Ziwei Liu\textsuperscript{1, \Letter}}

\authorrunning{Y. Cao et al.}

\titlerunning{HSImul3R}

\institute{%
\textsuperscript{1} S-Lab, Nanyang Technological University
\hspace{2 mm}
\textsuperscript{2} ACE Robotics\\
\textsuperscript{3} Shanghai AI Laboratory\\
\textsuperscript{\Letter} Corresponding author \\
{\small \url{https://yukangcao.github.io/HSImul3R/}}}

\maketitle

\begin{figure}
  \vspace{-8 mm}
  \includegraphics[width=\linewidth]{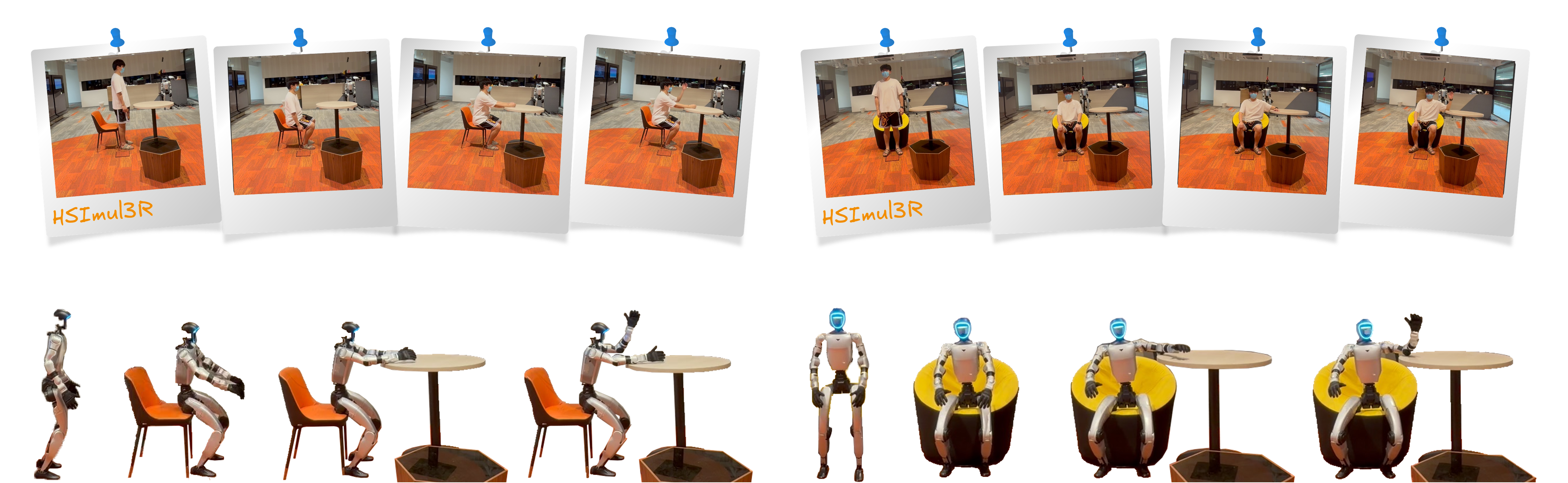}
  \caption{\textbf{Examples of real-world humanoid robotic deployment.} Given casual captures, our approach achieves simulation-ready 3D reconstruction of human–scene interactions by refining the human motions and scene geometry via a physically-grounded bi-directional optimization pipeline. Our optimized human motions can be seamlessly transferred and deployed in humanoid robotics.}
  \label{fig:g1_teaser}
  \vspace{-8 mm}
\end{figure}

\begin{abstract}
We present \textbf{\OM}\footnote{Pronunciation: \textipa{/ "sImjul@(r) /}}, a unified framework for simulation-ready 3D reconstruction of human-scene interactions (HSI) from casual captures, including sparse-view images and monocular videos.
Existing methods suffer from a \textit{perception-simulation gap}: visually plausible reconstructions often violate physical constraints, leading to instability in physics engines and failure in embodied AI applications.
To bridge this gap, we introduce a \textbf{physically-grounded bi-directional optimization pipeline} that treats the physics simulator as an active supervisor to jointly refine human dynamics and scene geometry.
In the forward direction, we employ \textit{Scene-targeted Reinforcement Learning} to optimize human motion under dual supervision of motion fidelity and contact stability.
In the reverse direction, we propose \textit{Direct Simulation Reward Optimization}, which leverages simulation feedback on gravitational stability and interaction success to refine scene geometry.
We further present \textbf{HSIBench}, a new benchmark with diverse objects and interaction scenarios.
Extensive experiments demonstrate that \OM produces the first stable, simulation-ready HSI reconstructions and can be directly deployed to real-world humanoid robots.


\end{abstract}
\section{Introduction}

Embodied artificial intelligence aims to integrate intelligent agents into daily life through physically grounded systems. 
Unlike disembodied models~\cite{xiu2023econ, cao2023sesdf, chen2025human3r, cai2025up2you} limited to virtual domains, embodied AI~\cite{ze2025twist, DBLP:journals/corr/abs-2601-22153, ze2025gmr, yin2025visualmimic} learns transferable motions that enable perception, reasoning, and action in real-world environments. 
A key challenge is modeling humanoid–scene interactions, requiring understanding of human motion, spatial layouts, and interaction stability. 
Reconstructing human–scene interactions (HSI)~\cite{bhatnagar2022behave, xie2025cari4d, lu2025humoto} from images or videos provides high-fidelity supervision and enables scalable, simulation-ready datasets, helping bridge passive observation and active robotic deployment.

Current methods suffer from a \textit{perception–simulation gap}, where visually plausible reconstructions violate physical constraints and fail in embodied AI applications.
This gap largely stems from the fragmented modeling of humans and environments, as existing approaches rarely capture their explicit physical coupling and instead fall into three separate directions:
\textbf{1) 3D scene reconstruction}
(\eg, NeRF~\cite{DBLP:conf/eccv/MildenhallSTBRN20}, Gaussian Splatting~\cite{DBLP:journals/tog/KerblKLD23}, DUSt3R~\cite{DBLP:conf/cvpr/Wang0CCR24}), which prioritizes environment geometry while largely ignoring human dynamics.
\textbf{2) Human motion estimation}~\cite{DBLP:conf/nips/CaiYZWSYPMZZLYL23, DBLP:journals/pami/TianZLW23, DBLP:conf/cvpr/PavlakosSRKFM24, DBLP:journals/pami/LiBXCYL25}, which achieves robustness under occlusion but reconstructs motion in isolation, without modeling physical contact or environmental constraints.
\textbf{3) Interaction modeling}~\cite{DBLP:conf/cvpr/YangL0LXLL22, DBLP:conf/cvpr/JiangZLMWCLZ024, DBLP:conf/cvpr/0009MGCPSJDXPWE24, DBLP:conf/cvpr/FanPK0KBH24, DBLP:conf/cvpr/PanYDWH0K025}, typically based on SMPL-driven HSI datasets~\cite{DBLP:conf/cvpr/BhatnagarX0STP22, DBLP:conf/iccv/JiangLCCZ0W0H23, DBLP:conf/iccv/LuHBHZ25} that remain limited in scale, diversity, and physical validation.
Recent unified frameworks (\eg, HOSNeRF~\cite{DBLP:conf/iccv/LiuCYXKSQS23}, HSfM~\cite{DBLP:conf/cvpr/MullerCZYMK25}) attempt joint modeling but optimize mainly in the 2D image space, prioritizing visual alignment over geometric and physical validity.
Consequently, the resulting reconstructions lack metric and contact fidelity, making them unsuitable for simulation and preserving the gap between visual realism and embodied deployment.

To close this gap, we introduce \OM, a simulation-ready Human–Scene Interaction 3D reconstruction framework that formulates reconstruction as a bi-directional physics-aware optimization problem.
A physics simulator acts as an active supervisor, enabling closed-loop refinement between human motion and scene geometry.
\OM operates along two complementary directions.
\textbf{Forward optimization} refines human motion under fixed scene geometry. 
After establishing metric-consistent human–scene alignment with structural priors from image-to-3D generative models, we integrate the reconstruction into the simulator and perform scene-targeted reinforcement learning. 
Motion is optimized using physically grounded signals, including keypoint tracking consistency and geometric contact constraints, improving interaction stability.
\textbf{Reverse optimization} refines scene geometry under physically validated motion. 
To address instability caused by structurally deficient geometry, we introduce Direct Simulation Reward Optimization (DSRO), which leverages simulator-derived rewards to enhance gravitational stability and interaction feasibility.

\begin{figure*}[!t]
  \centering
  \vspace{-8 mm}
  \includegraphics[width=\linewidth]{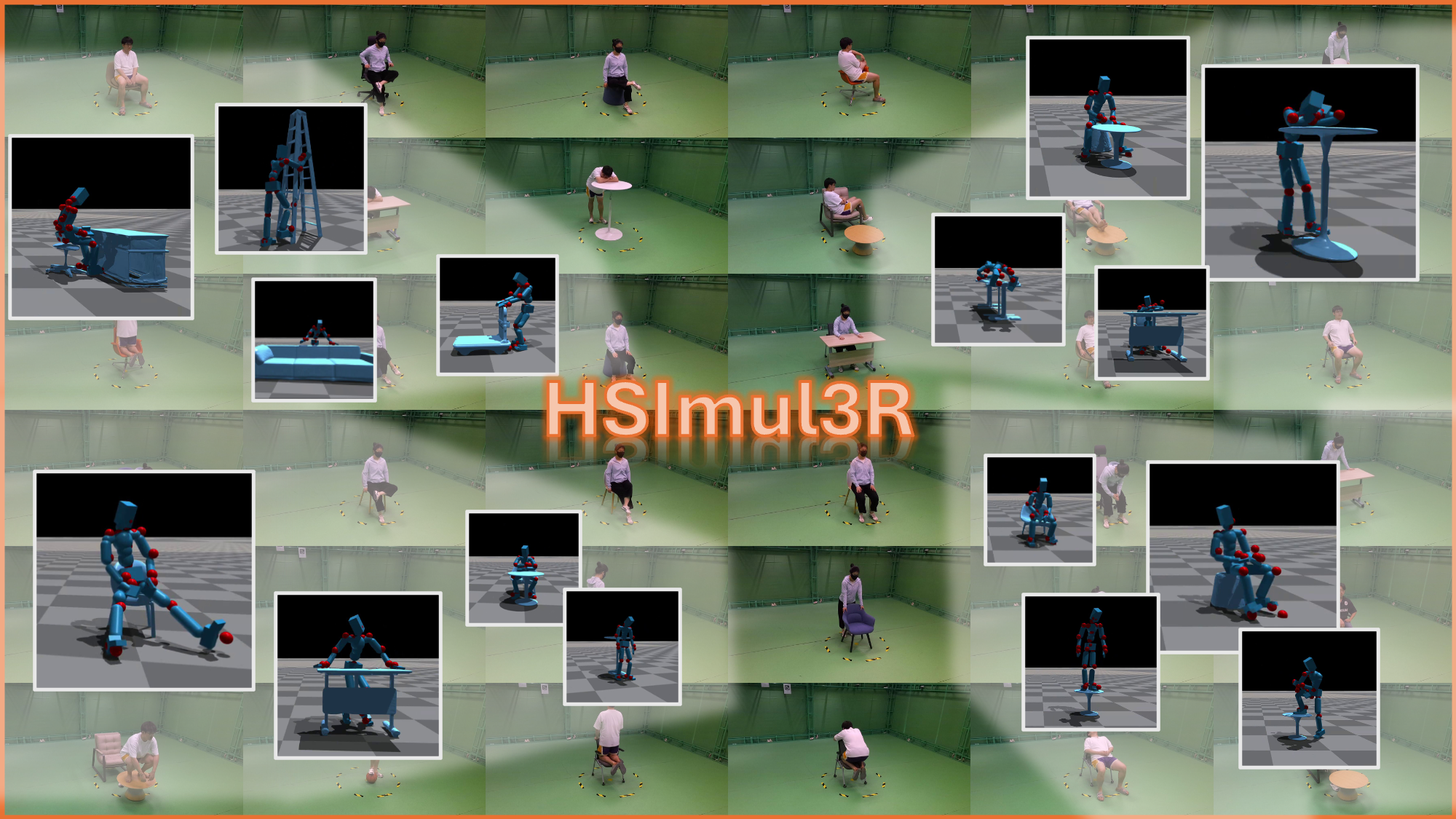}
  \vspace{-4 mm}
  \caption{\textbf{Examples of HSIBench and corresponding simulation results by \OM.} Our approach enables simulation-ready 3D reconstruction of human–scene interactions from casual captures. In addition, we collect HSIBench, a dataset comprising 16-view synchronized captures of diverse human–scene interactions, covering a wide range of scene objects, human subjects, and motions.}
  \label{fig:teaser}
  \vspace{-4 mm}
\end{figure*}
To support the training and benchmarking of this framework, we collect \textbf{HSIBench}, a new dataset comprising 19 objects and over 50 motion sequences recorded by two male and one female participants, totaling 300 unique interaction instances. An overview of HSIBench and simulation results of our method is provided in Fig.~\ref{fig:teaser}.

We conduct extensive experiments to evaluate \OM against state-of-the-art baselines in terms of simulation stability, post-simulation human motion quality, and improvements in image-to-3D generation through DSRO fine-tuning. 
Experimental results demonstrate that \OM is the first approach to achieve stable, simulation-ready reconstructions of human–scene interactions, offering robust performance across diverse scenarios and significantly outperforming existing techniques.
Finally, we demonstrate the real-world utility of our framework by (1) retargeting the refined motions to a Unitree humanoid robot, and (2) training a whole-body motion tracking policy for physical deployment. Examples of real-world deployment are presented in Fig.~\ref{fig:g1_teaser}.

\section{Related Works}

\paragraph{3D Scene Reconstruction}
Early approaches are dominated by geometry-based methods, such as structure-from-motion~\cite{DBLP:conf/cvpr/SchonbergerF16} and multi-view stereo~\cite{DBLP:conf/cvpr/SeitzCDSS06}, which estimate camera poses and dense geometry from multiple views. 
With the rise of deep learning, data-driven approaches emerge, including monocular depth prediction~\cite{DBLP:conf/cvpr/YangKHXFZ24,DBLP:conf/nips/YangKH0XFZ24} and learning-based multi-view stereo~\cite{DBLP:conf/cvpr/HuangMKAH18}, enabling reconstruction from sparse or unstructured imagery. 
Other works adopt explicit 3D representations such as voxels~\cite{DBLP:conf/cvpr/SongYZCSF17, DBLP:journals/ijcv/LiuXZYJNT25}, point clouds~\cite{DBLP:conf/cvpr/DaiCSHFN17, DBLP:conf/eccv/XieYZMZS20}, and meshes~\cite{DBLP:conf/cvpr/NieHGZCZ20}, often optimized through differentiable rendering.
More recently, implicit neural representations, such as signed distance functions~\cite{DBLP:conf/cvpr/ParkFSNL19}, occupancy fields~\cite{DBLP:conf/iclr/BianKXP0025}, neural radiance fields~\cite{DBLP:conf/iccv/0002JXX0L023, DBLP:conf/cvpr/Xie0H024}, and explicit but differentiable formulations like 3D Gaussian Splatting~\cite{DBLP:journals/tog/KerblKLD23, DBLP:conf/cvpr/Xie0H025}, become central to high-quality scene modeling.
Beyond static reconstruction, dynamic scene modeling~\cite{DBLP:conf/eccv/YanLZWSZLZP24, DBLP:journals/pami/XieCHL25} expands these methods to time-varying environments. 
In parallel, recent works such as Dust3R~\cite{DBLP:conf/cvpr/Wang0CCR24} and VGGT~\cite{DBLP:conf/cvpr/WangCKV0N25} introduce pre-trained transformers that enable end-to-end 3D reconstruction directly from uncalibrated and unlocalized images, eliminating the need for expensive post-optimization.

\paragraph{Physically-sounded Modeling}
Recent works have sought to embed physical soundness into modeling, which can be broadly categorized into three paradigms. 
Physics-constrained and physics-integrated generation methods unify simulation and content creation by leveraging simulation-derived losses or physical priors. 
For example, PhyRecon~\cite{DBLP:conf/nips/NiCJJW0L0Z024} ensures stable scene reconstruction, Atlas3D~\cite{DBLP:conf/nips/ChenXZLG0WJ24} and BrickGPT~\cite{DBLP:conf/iccv/PunDLRLZ25} produce self-supporting structures, and DSO~\cite{DBLP:conf/iccv/LiZRV25} or PhysDeepSDF~\cite{DBLP:conf/cvpr/MezghanniBBO22} align generators with simulation feedback. 
PhysGaussian~\cite{DBLP:conf/cvpr/XieZQLF0J24} evolves Gaussian splats via continuum mechanics, while PhyCAGE~\cite{DBLP:preprint/arxiv/2411-18548}, VR-GS~\cite{DBLP:conf/siggraph/JiangYXLFWLLG0J24}, and GASP~\cite{DBLP:preprint/arxiv/2409-05819} optimize assets through MPM; PAC/iPAC-NeRF~\cite{DBLP:conf/iclr/LiQCJLJG23, DBLP:conf/cvpr/Kaneko24} jointly learn geometry and physical parameters to bridge reconstruction and simulation. 
This approach also extends to interactive contexts: PhyScene~\cite{DBLP:conf/cvpr/YangJZH24} generates simulation-ready environments, PhysPart~\cite{DBLP:preprint/arxiv/2408-13724} models functional parts for robotics and fabrication, and DreMa~\cite{DBLP:conf/iclr/BarcellonaZAPGG25} produces manipulable, physics-grounded world models.

\paragraph{Human Simulation Imitating}
Recent advances in physics‑based humanoid simulation fall into three directions. 
Robust motion imitation builds on RL frameworks such as DeepMimic~\cite{DBLP:journals/tog/PengALP18} and AMP~\cite{DBLP:journals/tog/PengMALK21}, extended by PHC~\cite{DBLP:conf/iccv/0002CWKX23} for long‑horizon resilience and DiffMimic~\cite{DBLP:conf/iclr/RenYC0P023} with differentiable physics. 
More recent methods leverage human demonstrations for adaptive whole‑body imitation, including locomotion and manipulation, as in HumanPlus~\cite{DBLP:conf/corl/FuZWWF24}, TWIST~\cite{DBLP:preprint/arxiv/2505-02833}, and SFV~\cite{peng2018sfv}. 
Generalizable control is advanced by PULSE~\cite{DBLP:conf/iclr/0002CMWHKX24}, which provides compact latent spaces for versatile skills, HOVER~\cite{DBLP:preprint/arxiv/2410-21229}, which unifies multiple control modes, and diffusion‑based frameworks such as CLoSD~\cite{DBLP:conf/iclr/TevetRCR0PBP25} and InsActor~\cite{DBLP:conf/nips/RenZYMPL23}, which integrate generative planning with physics‑based execution for multi‑task behaviors. 
Interactive skills cover dynamic human‑object interactions and complex benchmarks: PhysHOI~\cite{DBLP:preprint/arxiv/2312-04393} and OmniGrasp~\cite{DBLP:conf/nips/0002CCWKX24} enable dexterous manipulation, SMPLOlympics~\cite{DBLP:journals/corr/abs-2407-00187} and HumanoidOlympics~\cite{DBLP:preprint/arxiv/2407-00187} provide sports environments, Half‑Physics~\cite{DBLP:journals/corr/abs-2507-23778} bridges kinematic avatars with physics, ImDy~\cite{DBLP:conf/iclr/0002LLH0L25} exploits imitation‑driven simulation, ASAP~\cite{DBLP:conf/rss/HeGXZWWLHSPYQKHFZLS25} improves fidelity by aligning dynamics with demonstration trajectories, BeyondMimic~\cite{liao2025beyondmimic} learns a motion tracking policy that could be deployed into humanoid robot, and VideoMimic~\cite{allshire2025visual} enables learning such policies from as little as a single monocular video.

\section{Our Approach}
\label{sec:method}

\begin{figure*}[!t]
  \centering
  \includegraphics[width=\linewidth]{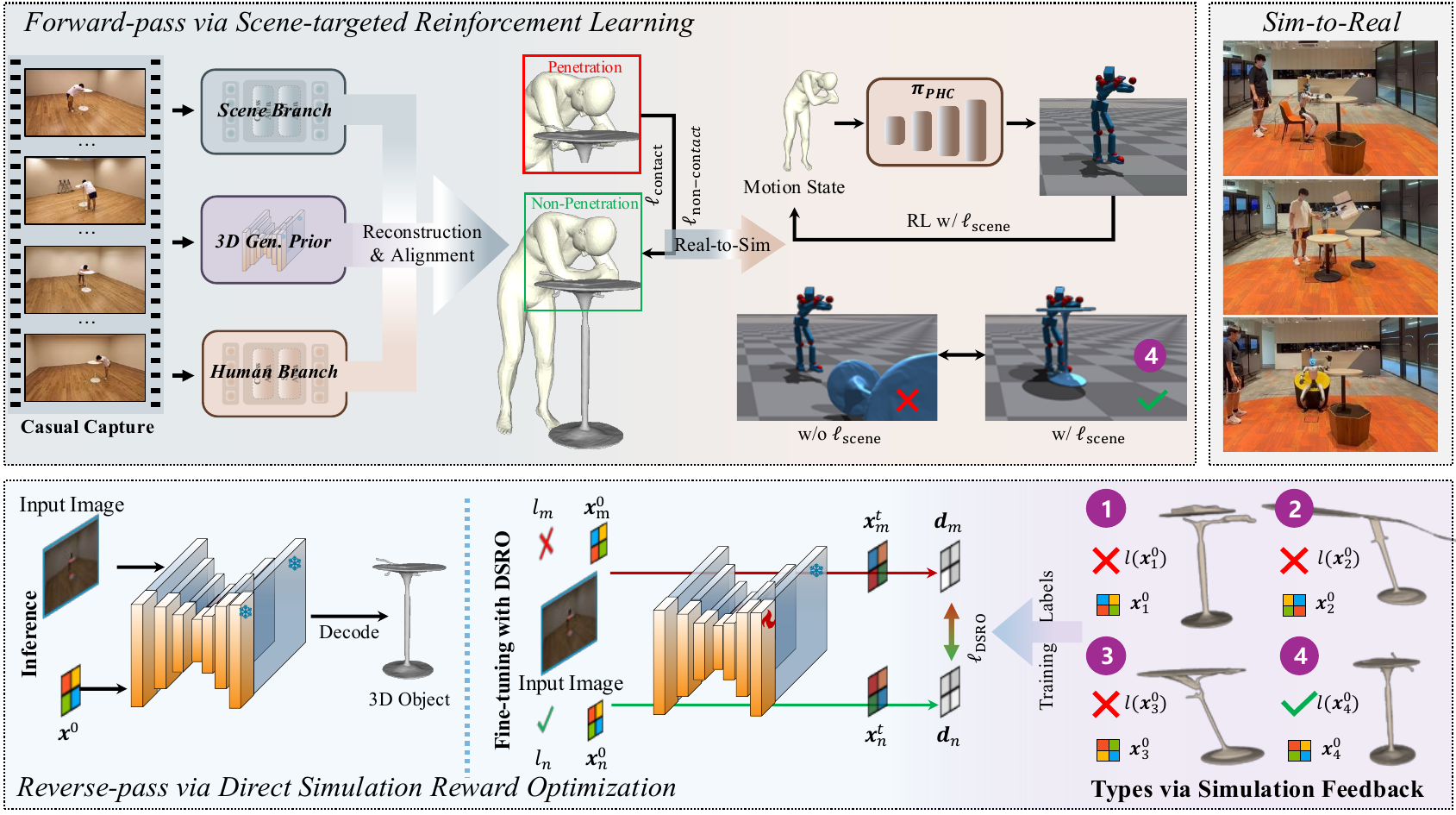}
  \caption{\textbf{Overview of \OM.} Given casual captures as inputs, we achieve simulation-ready reconstruction of human–scene interactions via a physics-in-the-loop optimization pipeline. We first propose to inject an 3D explicit generative prior into the reconstruction pipeline to achieve better alignment between human and scene. Then, \textbf{(1)} in the forward-pass, we propose a scene-targeted reinforcement learning that optimize the human motion to achieve interaction stability within the simulator, \textbf{(2)} in the reverse-pass, we introduce a direct simulation reward optimization (DSRO) to refine the scene geometry via simulation feedback regarding the stability. Specifically, we define the 4 types regarding the feedback. Type 1: objects not stabilizing under gravity; Type 2: objects failing to stabilize during human interaction; Type 3: objects stabilizing but without meaningful interaction; Type 4: objects with stable interaction.}
  \label{fig:pipeline}
\end{figure*}
As illustrated in Fig.~\ref{fig:pipeline}, \OM can reconstruct simulation-ready human-scene-interactions (HSI) from casual captures. 
To achieve this, we first reconstruct both human motion and scene geometry, subsequently aligning them through an explicit 3D structural prior derived from image-to-3D generative models~\cite{DBLP:conf/cvpr/HuangGAY0ZLLCS25} (Sec.~\ref{subsec:recon}).
Following this reconstruction, we introduce a physically-grounded bi-directional optimization pipeline. 
This process consists of a forward-pass optimization, which employs a proposed scene-targeted reinforcement learning scheme to refine human motions (Sec.~\ref{subsec:simulation}), followed by a reverse-pass optimization that leverages simulator feedback regarding physical stability to rectify the structural correctness of the scene geometry (Sec.~\ref{subsec:finetune}).
For simplicity of the illustration, we first focus on the setting of $J=4$ uncalibrated sparse-view inputs in the following sections and then discuss its extension to monocular videos in Sec.~\ref{subsec:video}.
Preliminaries underlying our methodology are provided in Sec.~\ref{subsec:pre}.

\subsection{Preliminaries}
\label{subsec:pre}

\paragraph{DUSt3R} 
Recently, DUSt3R~\cite{DBLP:conf/cvpr/Wang0CCR24} introduced a framework for 3D reconstruction that regresses point maps and employs a global alignment strategy to jointly predict depth maps and camera poses. Specifically, given a set of input images $\mathbf{I} = {I_0, I_1, ..., I_J}$, DUSt3R applies a ViT-based network that takes a pair of image frames $I_n, I_m$ $(n, m \in [0, J])$ to estimate the corresponding point maps $P^{e}_n, P^{e}_m \in \mathbb{R}^{H \times W \times 3}$ with respect to the coordinate system of frame $n$, along with confidence maps $C^{e}_n, C^{e}_m \in \mathbb{R}^{H \times W}$. Here, $e = (m, n)$ denotes the selected image pair. Aggregating point maps and confidence maps across selected pairs, DUSt3R builds a connectivity graph $\mathcal{G}(\mathcal{V}, \mathcal{E})$, where $\mathcal{V}$ corresponds to the $N$ images and $\mathcal{E}$ to the chosen image pairs $e$.

After collecting all pairwise point maps, DUSt3R performs a global alignment optimization to recover the depth maps $\mathbf{D} = {D_0, D_1, ..., D_J}$ and camera poses ${\pi_0, \pi_1, ..., \pi_J}$:
\begin{equation}
    \arg\min_{D, \pi, \sigma}\sum_{e\in\mathcal{E}}\sum_{n\in e} C_n^e||D_n - \sigma_e \cdot F_e(\pi_n, P_n^e)||_2^2,
\end{equation}
where $\sigma = {\sigma_e, e \in \mathcal{E}}$ denotes the edge-wise scale factors, and $F_e(\pi_n, P_n^e)$ projects the predicted point map $P_n^e$ to view $n$ under camera pose $\pi_n$ to produce the corresponding depth. This objective enforces geometric alignment across the input frame pairs, ensuring cross-view consistency in the estimated depth maps after the optimization. However, DUSt3R struggles with human subjects and frequently produces reconstruction artifacts such as unrealistic/incorrect scene structures and non-watertight topologies. Such defects make the reconstructed environments unreliable for stable simulation and downstream embodied AI tasks.

\subsection{Human-Scene Interaction Reconstruction and Alignment}
\label{subsec:recon}
The first stage of our pipeline involves the independent reconstruction of the static scene geometry and the dynamic human motion from uncalibrated captures. 
We adopt DUSt3R to recover the 3D structure of the environment.
For human motion estimation, we first utilize SAM2~\cite{DBLP:conf/iclr/RaviGHHR0KRRGMP25} to detect and associate individuals across frames, generating precise masks and identity tracks.
Following this, we employ 4DHumans~\cite{goel2023humans} and ViTPose~\cite{DBLP:conf/nips/XuZZT22} to extract the initial 3D SMPL-based motion sequences and 2D keypoints ($J_{2D}$), respectively. 
As the initial human and scene reconstructions may reside in disparate coordinate spaces, we perform a joint optimization to unify them~\cite{DBLP:conf/cvpr/MullerCZYMK25}. 
This is achieved through: (1) human-centric bundle adjustment guided by the 2D keypoints $J_{2D}$, and (2) global human-scene alignment, which minimizes the reprojection error between the ViTPose-detected keypoints and the projected 3D SMPL joints to ensure spatial consistency.

\paragraph{Alignment via Explicit 3D Structural Prior}
Despite the initial alignment listed above, two critical issues often persist: (1) the reconstructed scene geometry frequently contains structural artifacts, such as disconnected components, missing surfaces, or non-watertight topologies; and (2) the human-scene alignment relies solely on 2D projection-based supervision, which lacks 3D geometric awareness and is vulnerable to occlusions. These deficiencies inevitably lead to physical instability and drifting within the physics simulator. To resolve these challenges, we leverage 3D structural priors from pre-trained generative models to rectify the scene's geometry and enforce more robust interaction constraints.

Concretely, for each object within the scene, we automatically identify the input image $I_n, n \in [1, J]$ in which the object is most prominently featured and employ SAM~\cite{kirillov2023segment} to extract its segmentation mask. This view is then processed by a pre-trained image-to-3D generative model~\cite{DBLP:conf/cvpr/HuangGAY0ZLLCS25} to synthesize a high-fidelity 3D representation with better structural accuracy:
\begin{equation}
    \mathcal{R}_{\text{scene}} := \{\texttt{MIDI}(I_n[M_i]), i\in [0, O]\},
\end{equation}
where $\mathcal{R}_{\text{scene}}$ denotes the refined 3D scene and $O$ is the total number of objects. Note that our framework is flexible and allows for the usage of alternative or future more advanced models as they become available.

Thanks to the injection of 3D structural priors, we are now able to refine the human-scene alignment with 3D explicit constraints. This process is essential because penetration artifacts are particularly problematic in simulation: even minor inconsistencies in 3D space can manifest as severe collisions between body parts and objects, ultimately leading to unstable or failed simulations. To this end, we propose to optimize the position of the recovered human and generated objects~\cite{}. Specifically, if the object and human are not in contact, we optimize their positions via:
\begin{align}
    \ell_{\text{non-contact}} &= \frac{1}{|H_p|}\cdot \sum_{1 \leq j \leq N_o} ||\mu_i^h - \mu_j^o||_2 \nonumber \\
    &+ \frac{1}{N_o} \cdot \sum_{j=1}^{N_o} \min_{i\in H_p} ||\mu_j^o - \mu_i^h||_2,
\end{align}
where $H_P$ denotes the human body part closest to the object, and $N_o$ is the number of vertices on the object, and $\mu_j^o$ and $\mu_i^h$ represent the 3D positions of object and human vertices, respectively. When the object is in contact with the human, we instead apply:
\begin{equation}
    \ell_{\text{contact}} = \frac{1}{|H_p|} \cdot \sum_{i \in H_p} \text{max}(0, -\delta(\mu_i^h)),
\end{equation}
where $\delta(\cdot)$ denotes the signed distance function, measuring the penetration depth of the human vertex $\mu_i^h$ relative to the object surface.

\subsection{Forward-Pass: Scene-Targeted Motion Optimizatiton}
\label{subsec:simulation}
Following the initial 3D reconstruction of the human-scene interaction, the next step is to ensure stable dynamics within a physics simulator~\cite{DBLP:conf/nips/MakoviychukWGLS21}. 
A direct approach is to employ motion tracking techniques~\cite{DBLP:conf/iccv/0002CWKX23} to retarget the reconstructed human poses onto a humanoid robot.
However, directly simulating the raw reconstructions often fails to yield stable interactions (see Fig.~\ref{fig:qualitative_sim}). In many cases, the humanoid inadvertently displaces nearby objects, leaving them separated from the body and resting independently on the ground. 
This instability occurs because conventional 3D reconstructions do not account for gravity and interaction forces to verify if poses and object placements are physically realizable.

To address this issue, we introduce a scene-targeted supervision signal to reinforcement-learning-based motion tracking~\cite{DBLP:conf/iccv/0002CWKX23}.
Specifically, we propose an objective that enforces spatial proximity between the humanoid and relevant scene objects, encouraging physically plausible contact during simulation.
This loss is defined as the average Euclidean distance between human contact keypoints $k_j^h$ and their corresponding nearest object surface points $\mu_i^o$.
\begin{equation}
\label{eq:scene-targeted}
    \ell_{\text{scene}} = 
    \frac{1}{N_{\text{contact}} \cdot N_{\text{surf}}} \cdot
    \sum_{j=1}^{N_{\text{contact}}} 
    \sum_{i=1}^{N_{\text{surf}}} 
    \lVert \mu_i^o - k_j^h \rVert_2^2,
\end{equation}
where $N_{\text{contact}}$ is the number of contacts between the human and scene objects, and $N_{\text{surf}}$ denotes the number of sampled object surface points within the local contact region.

\subsection{Reverse-Pass: Simulator-Guided Object Refinement}
\label{subsec:finetune}
Nonetheless, even our forward-pass with scene-targeted reinforcement learning could enhance the simulation stability, we may still observe unsatisfactory stability ratios (see Tab.~\ref{tab:hsi}). 
As presented in Fig.~\ref{fig:qualitative_obj}, we observe that this problem largely stems from the inconsistent quality of our explicit 3D generative prior, for two main reasons: (1) generated objects often contain structural defects, especially in slender geometries. For example, tables or chairs may be missing legs, making them unstable in the simulator even without interaction; and (2) severe occlusion by the human in the input images, which frequently happens, often results in generated objects exhibiting artifacts, such as surface distortions or unwanted bumps. Together, these limitations make it difficult for the humanoid to establish stable and physically plausible contact during simulation.

\paragraph{Direct Simulation Reward Optimization}
Inspired by DSO~\cite{DBLP:conf/iccv/LiZRV25}, we address this issue by introducing Direct Simulation Reward Optimization (DSRO), a novel approach that leverages physics-based simulation feedback as a supervision signal for refining 3D explicit object generation. Unlike methods that rely on human annotations or 3D ground truth, DSRO directly exploits the outcome of the simulation to assess the physical plausibility of generated objects and their interactions with humans.

Formally, we define the DSRO objective as:
\begin{equation}
\begin{split}
    \ell_{\text{DSRO}} &= -T\cdot \mathbb{E}_{I\sim\mathcal{I}, x_0\sim\mathcal{X}_I, t\sim \mathcal{\mu}(0, T), x_t\sim q(x_t|x_0)} \\
    & [w(t) \cdot (1 - 2 \cdot l(x_0))||\epsilon - \epsilon_{\theta}(x_t, t)||_2^2]],
\end{split}
\end{equation}
where $I$ denotes an image sampled from the training dataset $\mathcal{I}$, $\mathcal{X}$ corresponds to its generated 3D explicit object, and $l(\cdot)$ encodes the stability feedback obtained from simulation.
Crucially, we define the stability $l(x_0)$ based on both gravitional stability and interaction stability:
\begin{equation}
    l(x_0) = 
    \begin{cases}
        1, & \quad \text{if stable} \\
        0, & \quad \text{otherwise},
    \end{cases}
\end{equation}
where stability is determined according to three criteria: (1) the object must remain upright and physically stable under gravity within the simulator, (2) it must achieve a stable final state for the reconstructed scene, and (3) the interaction must involve actual contact rather than the object resting independently on the ground.

\paragraph{HSIBench} 
To support the training and benchmarking of this framework, we construct a dedicated benchmark dataset, HSIBench, tailored for human–scene interaction (HSI). The dataset is built by systematically capturing interaction scenarios involving three volunteers (two male and one female) engaging with a diverse set of objects, including eight chairs, three tables, and three sofas. In total, we record 300 distinct HSI cases, with each case captured from 16 different viewpoints to provide rich multi-view supervision. Representative examples are illustrated in the Appendix. We employ multi-view 2DGS reconstruction~\cite{Huang2DGS2024} and SMPL estimation to respectively derive pseudo ground truth for object geometry and human motion, for our quantitative evaluations.
For every captured case, we run our full reconstruction and simulation pipeline, as described in Sec.~\ref{subsec:recon} and Sec.~\ref{subsec:simulation}, 15 times under different random seeds for each view. This procedure ensures variability in the simulation outcomes and allows us to systematically collect the training signals needed for fine-tuning.

\subsection{Extension to Monocular Videos}
\label{subsec:video}

Thanks to the dynamic nature of the motion tracking techniques~\cite{DBLP:conf/iccv/0002CWKX23} employed in our forward-pass (Sec.~\ref{subsec:simulation}), our method can be easily extended to take monocular video as input to produce simulation-ready 4D reconstructions.
In this pipeline, we employ MegaSAM~\cite{li2025megasam} and TRAM~\cite{wang2024tram} for scene reconstruction and human motion estimation~\cite{allshire2025visual}, respectively.
We currently assume a static scene in which the human subject is the only moving entity. 
To achieve dynamic 3D alignment between the human and the scene geometry, we employ SAM2~\cite{DBLP:conf/iclr/RaviGHHR0KRRGMP25} to extract 2D bounding boxes for both the person and specific scene geometry (\eg, tables, chairs). 
This spatial knowledge allows us to preliminarily identify interactions and achieve accurate 3D alignment throughout the sequence.

\section{Experiments}
\label{sec:experiments}

\begin{table*}[t]
\centering
\caption{\textbf{Quantitative comparison regarding human-scene-interaction reconstruction and simulation.} We assess (1) post-simulation interaction stability, (2) human–scene penetration in the 3D reconstruction, and (3) changes in the human motions after simulation. Our method consistently outperforms the baseline method (HSfM) and different variants.}
\label{tab:hsi}
\begin{tabularx}{\linewidth}{X YYY c cc}
\toprule
  \multirow{2}*{Method} & 
  \multicolumn{3}{c}{Stability-HSI ($\%$) $\uparrow$} & 
  Scene Penetration & 
  \multicolumn{2}{c}{Human Motion Quality} \\
  \cmidrule(l){2-4} \cmidrule(l){5-5} \cmidrule(l){6-7} 
  {} & Easy & Medium &  Hard & 
  SP-3D ($\%$) $\downarrow$ & 
  W-MPJPE $\downarrow$ & PA-MPJPE $\downarrow$ \\
\midrule
  HSfM~\cite{DBLP:conf/cvpr/MullerCZYMK25}   & 10.52 & 4.50 & 2.66 & 69.51 & 5.02 & 2.79 \\  
  \textbf{V1}  & 13.96 & 8.81 & 4.17 & 77.12 & 6.18 & 3.20 \\
  \textbf{V2}    & 39.56 & 22.71 & 7.05 & - & 4.91  & 2.71 \\
  \textbf{V3} & 42.57 & 23.84 & 10.18 & - & 4.60 & 2.42 \\
  \textbf{V4} & 29.56 & 16.62 & 5.17 & - & 4.57 & 2.39 \\
\midrule
  Ours  & \textbf{53.68} & \textbf{30.56}  & \textbf{13.92} & \textbf{22.9} & \textbf{4.09} & \textbf{2.17}\\

\bottomrule
\end{tabularx}

\end{table*}
\begin{table*}[t]
\centering
\caption{\textbf{Quantitative comparison regarding image-to-3D generation quality.} Stability-HSI’ denotes the ratio of simulations in which the human–scene interaction remains stable, while ``Stability-Gravity''(SG) refers to placing the object alone in the simulator and evaluating whether it can stand stably under gravity. Note that we fine-tune DSO on pre-trained MIDI model for fair comparison.}
\label{tab:obj}
\begin{tabularx}{\linewidth}{l YYY Y Y c}
\toprule
  \multirow{2}*{Method} & 
  \multicolumn{3}{c}{Stability-HSI ($\%$) $\uparrow$} & 
  \multirow{2}* {SG ($\%$) $\uparrow$} & 
  \multirow{2}* {CD $\downarrow$} & 
  \multirow{2}* {F-Score $\uparrow$} \\
  \cmidrule(l){2-4}
   & {Easy} & {Medium} &  {Hard} &  &  & \\
\midrule
  MIDI~\cite{DBLP:conf/cvpr/HuangGAY0ZLLCS25}   & 29.56 & 16.62 & 5.17 & 79.19 & 0.198 & 81.95 \\
  DSO*~\cite{DBLP:conf/iccv/LiZRV25}  & 38.75 & 25.91 & 7.88 & 87.23 & 0.191 & 86.26 \\
\midrule
  Ours  & \textbf{53.68} & \textbf{30.56}  & \textbf{13.92} & \textbf{91.50} & \textbf{0.173} & \textbf{88.25}\\
\bottomrule
\end{tabularx}


\end{table*}

We evaluate \OM across three dimensions: reconstruction fidelity, simulation stability, and the impact of fine-tuning with the proposed DSRO. We also benchmark against existing methods and perform ablation studies to assess the contribution of each component.

\paragraph{Implementation Details}
Our approach is developed on top of HSfM~\cite{DBLP:conf/cvpr/MullerCZYMK25} and PHC~\cite{DBLP:conf/iccv/0002CWKX23}. For training, we adopt AdamW~\cite{loshchilov2017decoupled} as the optimizer and fine-tune the pre-trained MIDI model using LoRA~\cite{DBLP:conf/iclr/HuSWALWWC22}. Specifically, we set the LoRA rank to 64, use a batch size of 1, and a learning rate of $10^{-5}$. 
The model is trained for a total of 1800 steps on four NVIDIA A100 GPUs.

\begin{figure}[!t]
    \centering
    \includegraphics[width=\linewidth]{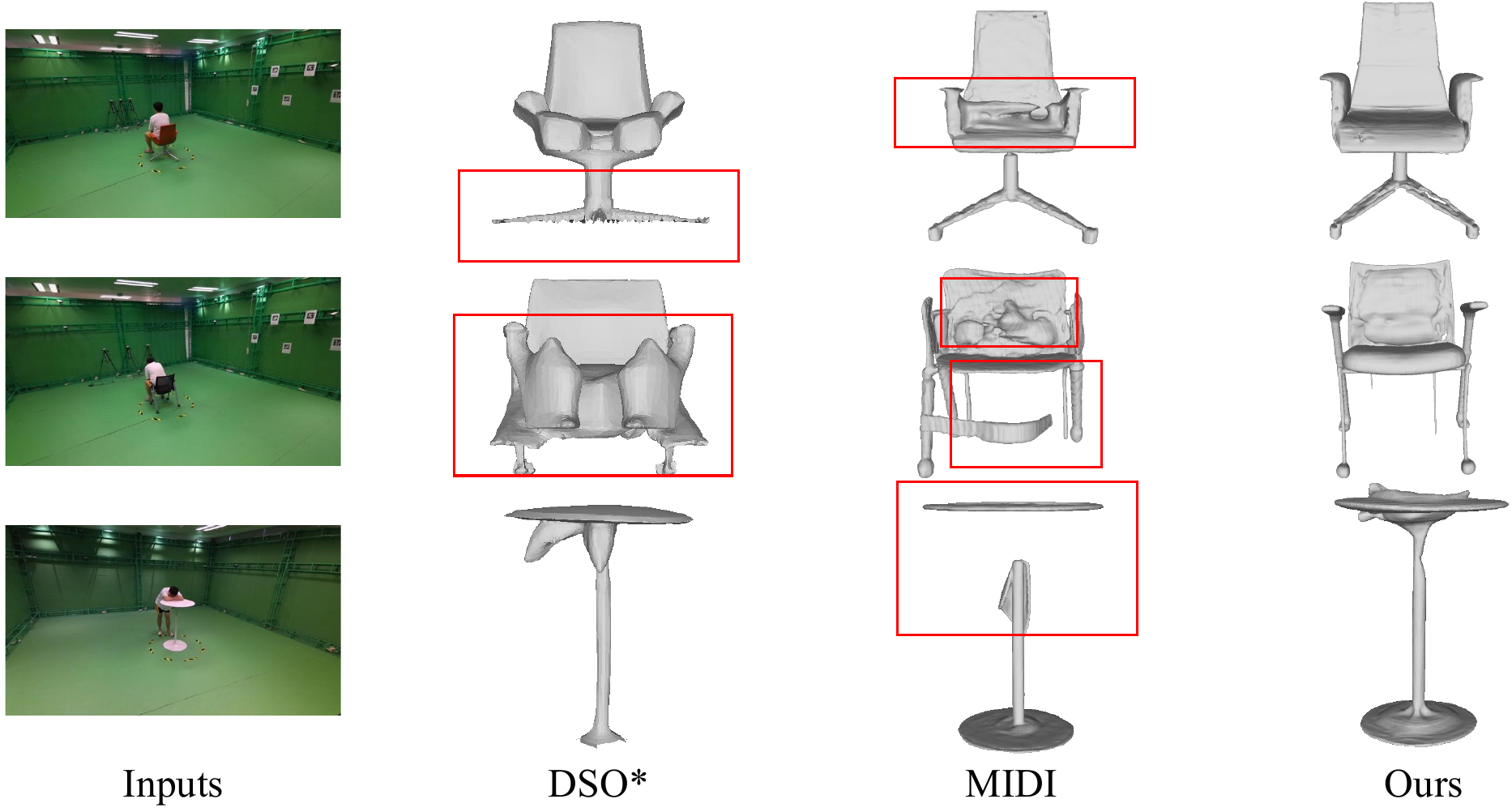}
    \caption{\textbf{Qualitative comparison regarding image-to-3D object reconstruction.} Our method enhances the object’s geometric structure while reducing surface ``bumps'' that may negatively impact human interaction.}
    \vspace{-4 mm}
    \label{fig:qualitative_obj}
\end{figure}
\paragraph{Baseline Methods}
Since our method presents the first approach for simulation-ready reconstruction of human–scene interactions from uncalibrated sparse-view inputs, we primarily compare its performance against HSfM~\cite{DBLP:conf/cvpr/MullerCZYMK25}, which is the first and only technique to reconstruct 3D HSI under sparse-view settings. 
Additionally, considering that there is no other dedicated method existing for this task, we further compare with various alternatives: 
\textbf{(V1)} a simple baseline that integrates HSfM with MIDI~\cite{DBLP:conf/cvpr/HuangGAY0ZLLCS25} and feeds the resulting reconstruction into the simulator;
\textbf{(V2)} using the reconstruction from Sec.~\ref{subsec:recon} directly in the simulator without applying the scene-targeted distance minimization of Eq.~\ref{eq:scene-targeted};
\textbf{(V3)} Replacing our object-surface distance computation with a center-point distance following CLoSD~\cite{DBLP:conf/iclr/TevetRCR0PBP25};
\textbf{(V4)} Obtain the simulated reconstruction directly via Sec.~\ref{subsec:recon} and Sec.~\ref{subsec:simulation} without fine-tuning the generative model using simulation feedback via the proposed DSRO.

We also compare with the MIDI~\cite{DBLP:conf/cvpr/HuangGAY0ZLLCS25} and DSO~\cite{DBLP:conf/iccv/LiZRV25} in terms of the geometric quality of the generated scene objects, as well as stability under both gravity-only and HSI scenarios. For fairness, we fine-tune DSO on the pre-trained MIDI model rather than its originally used Trellis~\cite{DBLP:conf/cvpr/XiangLXDWZC0Y25} model.

\paragraph{Evaluation Metrics}
We first evaluate the penetration ratio (SP-3D) in the reconstructed 3D HSI scenes. 
Next, we assess the stability of simulated human–scene interactions using the metric ``Stability-HSI'', which considers three factors: (1) object stability under gravity, (2) whether the HSI scene reaches a stable state in the simulator, and (3) whether the final state preserves meaningful human–scene interactions. Finally, we evaluate the quality of simulated human motion by extracting it from the final state and comparing it to the ground truth. 
Following HSfM, we report W-MPJPE for accuracy in the world coordinate system and PA-MPJPE for local pose precision.

For reconstructed 3D scene objects, we measure geometric quality using Chamfer Distance and F-Score, while physical plausibility is evaluated through ``Stability-HSI'' and ``Stability-Gravity''.

\paragraph{Evaluation Datasets}
We perform both quantitative and qualitative evaluations on our collected HSIBench dataset. To assess HSI simulation stability across different scenarios, we divide HSIBench into three levels of difficulty, \textit{i.e}, easy, medium, and hard, based on interaction complexity.

\begin{figure}[!t]
  \centering
  \includegraphics[width=\linewidth]{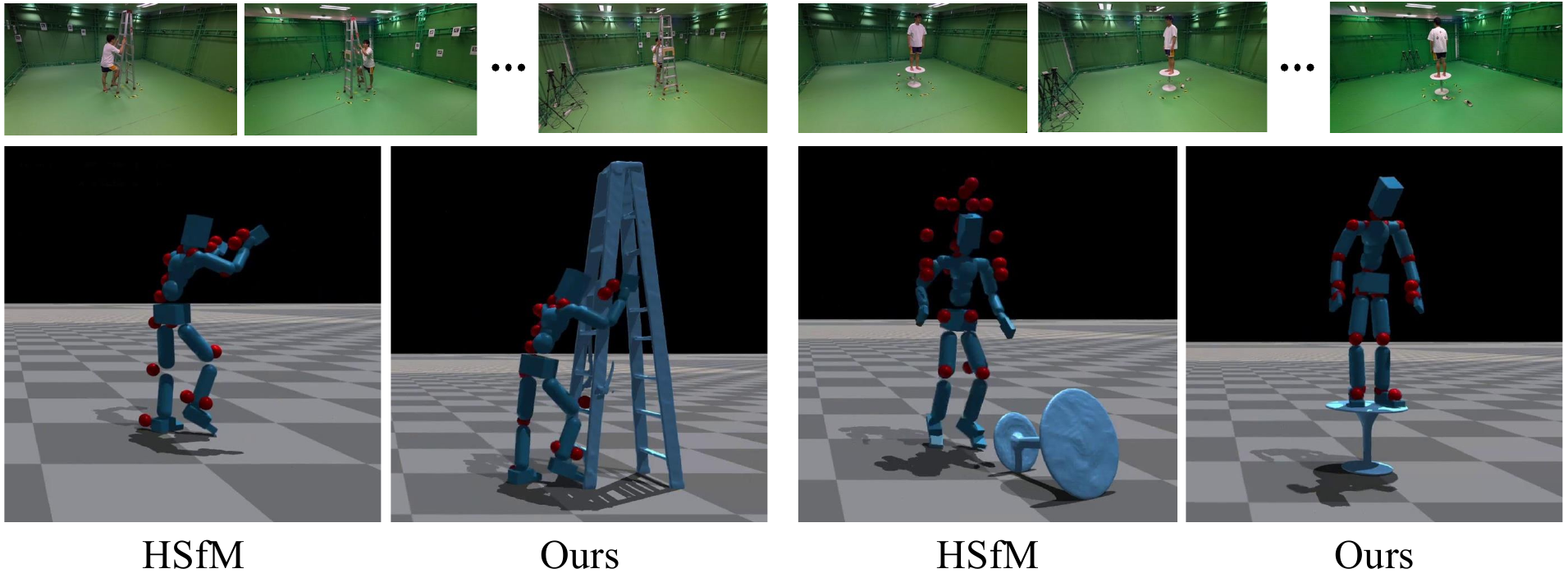}
  \caption{\textbf{Qualitative comparisons with HSfM~\cite{DBLP:conf/cvpr/MullerCZYMK25}.} Due to challenges such as (1) penetration issues and (2) inaccurate scene-object structures with geometric distortions, HSfM often struggles to achieve stable interactions in the simulator, frequently leading to unintended object displacement.
  }
  \label{fig:qualitative_sim}
  \vspace{-4 mm}
\end{figure}

\subsection{Results and Analysis}

\paragraph{Quantitative Evaluations}
We first present quantitative comparisons of HSI reconstruction and simulation quality in Tab.~\ref{tab:hsi}. As shown, our method significantly outperforms the only existing baseline, HSfM, as well as the ablated variants, across all evaluated metrics. This demonstrates both the overall effectiveness of our approach and the contribution of the proposed components. Note that V1, V2, and V3 do not report scene penetration percentages, as their 3D reconstruction is identical to that of V1. We then report quantitative results on image-to-3D generation quality in Tab.~\ref{tab:obj}. Importantly, our method (incorporating DSRO) achieves improved physical plausibility and interaction stability, along with superior geometric accuracy.

\paragraph{Qualitative Evaluations}
\textbf{(1)} In Fig.~\ref{fig:qualitative_sim}, we present the qualitative comparisons with HSfM. Specifically, we apply Poisson reconstruction to the point maps generated by HSfM and place the reconstructed objects into the simulator for evaluation. 
As shown, HSfM often fails to produce stable human–object interactions: the human frequently kicks objects away and ends up standing alone. In contrast, our method consistently achieves stable interaction states within the simulation.
\textbf{(2)} Fig.~\ref{fig:qualitative_obj} further compares our approach with DSO and MIDI in terms of image-to-3D reconstruction quality. Both baselines struggle to recover accurate structures and often introduce geometric distortions, which in turn lead to instability during simulation. 
By contrast, our DSRO fine-tuned model mitigates these issues, yielding more structurally faithful and stable reconstructions.

\begin{figure}[!t]
  \centering
  \includegraphics[width=\linewidth]{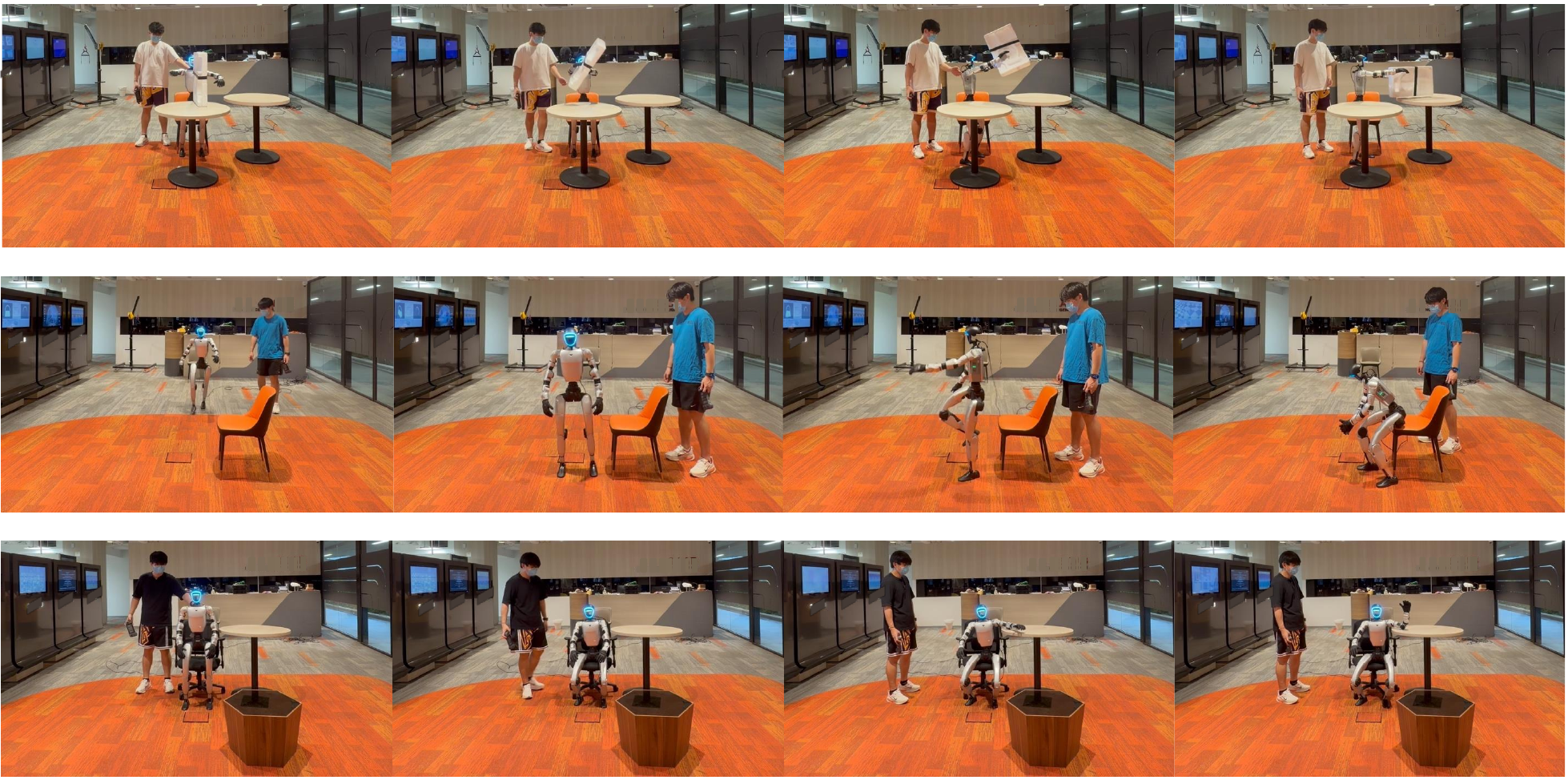}
  \caption{\textbf{Sim-to-Real results deployed on Unitree G1 humanoid robotics.} The refined human motions by our method could be successfully transferred and deployed on a real-world humanoid robot to conduct various human-scene-interaction scenarios.}
  \label{fig:g1_results}
  \vspace{-2 em}
\end{figure}

\paragraph{Real-world Robotics Deployment}
Building upon the refined human motions detailed in Sec.~\ref{subsec:simulation}, we utilize GMR~\cite{ze2025gmr, joao2025gmr} to retarget human trajectories onto the Unitree G1 humanoid robots.
These retargeted motions then serve as a prior for diffusion-guided reinforcement learning~\cite{liao2025beyondmimic}, which we employ to train a whole-body control policy within the IsaacGym simulator~\cite{makoviychuk2021isaac}.
This framework allows the agent to learn robust balancing by leveraging the generative priors of diffusion models during the RL training phase.
Once trained, the resulting control policy is deployed directly onto the physical G1 humanoid hardware via the Unitree SDK.
As illustrated in Fig.~\ref{fig:g1_results}, the successful deployment of the resulting policy on the physical Unitree G1 robot demonstrates that our refined motions facilitate robust robot-scene interactions.
This framework provides a scalable foundation for leveraging vast, cost-effective datasets from platforms like YouTube to augment existing training data for large-scale embodied AI models. Ultimately, our approach promotes the execution of complex robotic tasks and supports various downstream applications in autonomous robot-scene interaction.

\begin{table*}[t]
\centering
\caption[Quantitative analysis of number of input views.]{\textbf{Quantitative analysis of number of input views.} We quantitatively evaluate reconstruction quality and simulation stability under different numbers of input views. While additional views slightly improve motion quality, they don't have obvious impact over the interaction stability and increase scene penetration.}
\label{tab:views}
\begin{tabularx}{\linewidth}{l YYY c cc}

\toprule
  \multirow{2}*{Method} & 
  \multicolumn{3}{c}{Stability-HSI ($\%$) $\uparrow$} & 
  Scene Penetration & 
  \multicolumn{2}{c}{Human Motion Quality} \\
  \cmidrule(l){2-4} \cmidrule(l){5-5} \cmidrule(l){6-7} 
  {} & Easy & Medium &  Hard & 
  SP-3D ($\%$) $\downarrow$ & 
  W-MPJPE $\downarrow$ & PA-MPJPE $\downarrow$ \\
\midrule
16-view  & \textbf{55.16} & 29.51 & 13.59 & 21.81 & \textbf{4.01} & \textbf{1.99} \\
10-view  & 52.93 & \textbf{32.17} & 13.03 & \textbf{21.00} & 4.06  & 2.05 \\
4-view  & 53.68 & 30.56  & \textbf{13.92} & 22.90 & 4.09 & 2.17\\
\midrule[1pt]
\end{tabularx}

\end{table*}

\begin{figure}[!t]
  \centering
  \includegraphics[width=\linewidth]{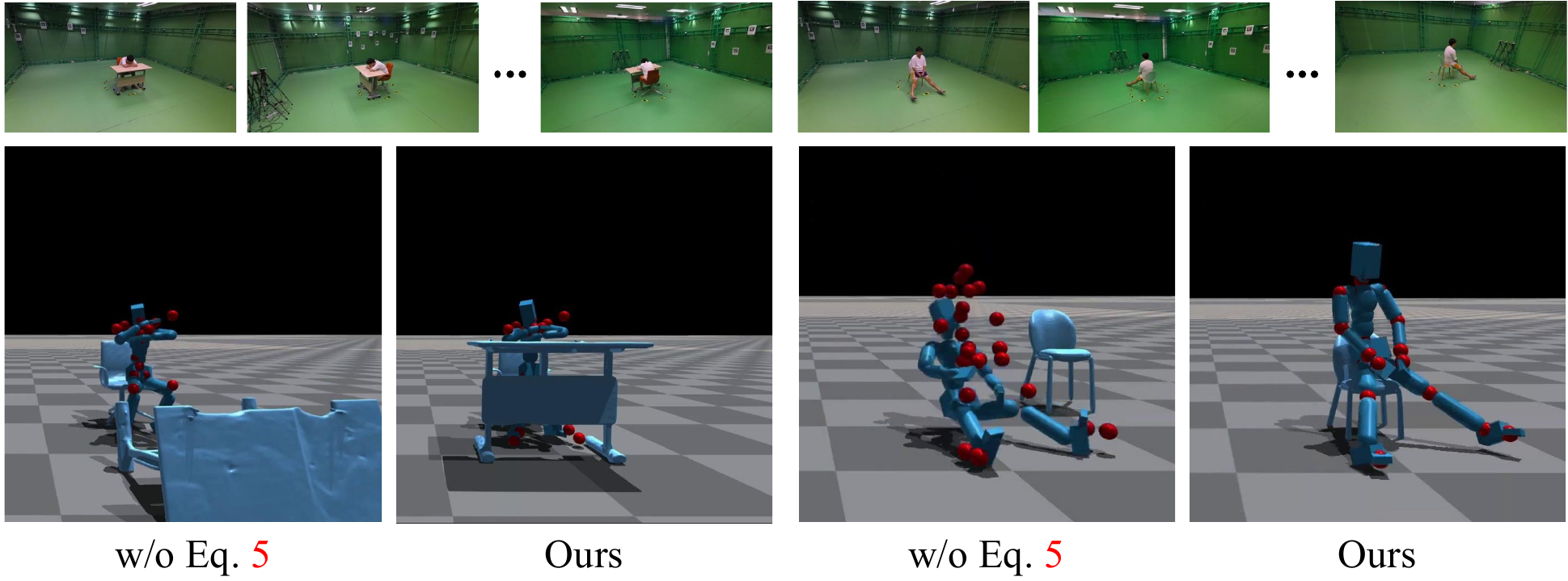}
  \caption{\textbf{Ablation studies on Eq.~\ref{eq:scene-targeted}.} Without the proposed scene-targeted RL, the simulation often results in unintended object displacement and fails to maintain stable interactions.}
  \label{fig:qualitative_eq5}
\end{figure}

\paragraph{Analysis of Scene-targeted Simulation}
Fig.~\ref{fig:qualitative_eq5} presents ablation studies on scene-targeted simulation loss defined in Eq.~\ref{eq:scene-targeted}. 
Results indicate that removing the distance-minimization term destabilizes the humanoid, leading to exaggerated motions and often kicking objects away.

\paragraph{Analysis of the Number of Input Views}
In Tab.~\ref{tab:views}, we analyze the effect of the number of input views. The results indicate that increasing the number of views leads to slight improvements in human motion quality. Interestingly, however, we notice that the number of views has little impact on penetration handling or the overall stability of the simulation.

\paragraph{Limitation and Failure Cases}
While \OM represents the first attempt at simulation-ready reconstruction of human–scene interactions, we acknowledge that our method has certain limitations: \textbf{(1)} the successful ratio is not very high, particularly in scenarios involving complex interactions or multiple objects (more than three);
\textbf{(2)} In many failure cases, the humanoid and objects tend to end up standing independently rather than engaging in meaningful interactions (see supplementary material); 
\textbf{(3)} Our fine-tuned image-to-3D model inevitably inherits biases from both the MIDI original training dataset and our collected HSIBench, which may constrain its generalizability to out-of-domain cases.

\section{Conclusion}

In this work, we introduced \OM, the first framework for reconstructing simulation-ready human–scene interactions from uncalibrated sparse views. Our approach incorporates a contact-aware interaction model to mitigate human–scene penetration issues in 3D reconstruction, a scene-targeted reinforcement learning strategy to promote stable interactions within the simulator, and a direct simulation reward optimization scheme that leverages simulation feedback to fine-tune the image-to-3D generative model, thereby improving simulation success rates. To support both training and evaluation, we collected the HSIBench dataset. Extensive experiments demonstrate that \OM achieves high-fidelity results, delivering both stable simulations and high-quality image-to-3D reconstructions, and outperforms existing state-of-the-art methods.

\clearpage
\bibliographystyle{splncs04}
\bibliography{references}


\end{document}